\documentclass[twoside]{article}
\usepackage{xcolor}
\usepackage{soul}
\usepackage[accepted]{aistats2024}

\usepackage{amsmath}
\usepackage{amsthm}

\usepackage[numbers]{natbib}

\setcitestyle{authoryear} 

\usepackage[utf8]{inputenc} 
\usepackage[T1]{fontenc}    
\usepackage{url}            
\usepackage{booktabs}       
\usepackage{amsfonts}       
\usepackage{nicefrac}       
\usepackage{microtype}      
\usepackage{xcolor}         

\usepackage{graphicx}
\usepackage[colorlinks=true, allcolors=blue]{hyperref}
\newtheorem{theorem}{Theorem}[section]

\newtheorem{definition}[theorem]{Definition}
\newtheorem{remark}[theorem]{Remark}

\newcommand{\Bern}{\text{Bern}}

\newcommand{\Pb}{\mathbb{P}}

\newcommand{\bitem}{\begin{itemize}}
\newcommand{\eitem}{\end{itemize}}
\newcommand{\benum}{\begin{enumerate}}
\newcommand{\eenum}{\end{enumerate}}
\newcommand{\beq}{\begin{equation}}
\newcommand{\eeq}{\end{equation}}
\newcommand{\beqs}{\begin{equation*}}
\newcommand{\eeqs}{\end{equation*}}
\newcommand{\bas}{\begin{align*}}
\newcommand{\eas}{\end{align*}}

\DeclareMathOperator*{\argmin}{arg\,min}

\begin{document}

\twocolumn[

\aistatstitle{FairRR: Pre-Processing for Group Fairness through Randomized Response}

\aistatsauthor{ Xianli Zeng* \And Joshua Ward* \And Guang Cheng}
\runningauthor{Xianli Zeng, Joshua Ward and Guang Cheng}

\aistatsaddress{ \texttt{xlzeng@wharton.upenn.edu}\\
NUS (Chongqing) Research Institute  \And \texttt{joshuaward@ucla.edu}\\
University of California\\ Los Angeles  \And \texttt{guangcheng@ucla.edu}\\
University of California\\
Los Angeles } 
]
\def\thefootnote{*}\footnotetext{These authors contributed equally to this work}\def\thefootnote{\arabic{footnote}}

\begin{abstract}
The increasing usage of machine learning models in consequential decision-making processes has spurred research into the fairness of these systems. While significant work has been done to study group fairness in the in-processing and post-processing setting, there has been little that theoretically connects these results to the pre-processing domain. This paper proposes that achieving group fairness in downstream models can be formulated as finding the optimal design matrix in which to modify a response variable in a Randomized Response framework. We show that measures of group fairness can be directly controlled for with optimal model utility, 
proposing a pre-processing algorithm called FairRR \footnote{All code for FairRR with corresponding experiments can be found at: \url{https://github.com/UCLA-Trustworthy-AI-Lab/FairRR}} that yields excellent downstream model utility and fairness.

\end{abstract}

\section{INTRODUCTION}

As the use of machine learning models becomes increasingly prevalent in decision-making processes, concerns about the fairness of algorithms have become more pressing. Case studies from various domains such as criminal justice, healthcare, and employment (\cite{FBC2016,corbett2018measure,JJSL2016, SEC2019}), have demonstrated that biased algorithms can perpetuate or even amplify discrimination against individuals and groups. In response, a variety of approaches have been developed to ensure fairness focusing on the pre-processing of data, the in-processing of models, or the post-processing of model predictions (\cite{ZWKT2013,CKYMR2016,CFDV2017,XWYZW2019,LV2019,CJG2019,DETR2018,CMJW2019,JL2019,CHS2020,zeng2022bayesoptimal}). This litany of methods extends across many metrics of fairness that can roughly be broken into two groups: group fairness (\cite{CFM2009,DHPT2012,HPS2016}) where fairness is defined as ensuring various types of statistical parity across distinct protected groups, and individual fairness (\cite{JKMR2016,PKG2019,RBMF2020}) which aims to provide nondiscriminatory predictions for similar individuals. 

In this paper, we focus on common group fairness criteria,
including demographic parity (\cite{CFM2009,KAS2012,CHS2020}), equality of opportunity (\cite{HPS2016,ZLM2018,CHS2020}), and predictive equality (\cite{CSFG2017}) adding to a larger family of diverse pre-processing methods in the supervised classification setting. In general, the goal of pre-processing is to modify the feature space of the original dataset such that when a classifier is trained on this altered data its output is fair. Strategies for this include transforming the data (\cite{FFMS2015,KJ2016,CFDV2017,JL2019}), fair representation learning (\cite{ZWKT2013,CKYMR2016,DETR2018,CMJW2019}) and fair generative models (\cite{XYZ2018,SHC2019,XYfairplus,RKSR2021}). These methods are convenient to apply, as they do not change the training procedure and are generally independent to downstream modelling tasks, allowing for the use of most classifiers. However, they often do not allow for the control of the exact fairness level, they do not always have full coverage of the variety of group fairness metrics in use, and they do not take advantage of recent results from the fair statistical learning literature.

Indeed, with such a variety of potential strategies to deploy in ensuring algorthims are fair, a robust body of literature has developed to answer the question of what exactly is the best theoretical classification strategy in terms of model utility and fairness. In the in-processing domain where fairness is achieved through the modification of a classifier itself, first \cite{CSFG2017} proved that, under several group fairness metrics, the fair Bayes-optimal classifiers are group-wise thresholding rules with unspecified thresholds. \cite{MW2018} related demographic parity and equality of opportunity to cost-sensitive risks and derived fair Bayes-optimal classifiers under these two fairness measures. Under the setting of perfect demographic parity and equality of opportunity, exact forms of fair Bayes-optimal classifiers were derived in \cite{CCH02019} and \cite{SC2021}, respectively. Finally, \cite{zeng2022bayesoptimal} and \cite{zeng2022fair} showed that in the general case, fair Bayes-optimal classifiers could be derived for any level of disparity in most definitions of group fairness which is an advantage in the applied setting if some level of unfairness is allowed for better model utility. This paper extends this line of research to the pre-processing area where our goal is to develop a method that allows for an explicit level of control of disparity in training data and to extend these theoretical results to create a unified framework for adjusting for disparity at every step of machine learning model development.


Thus, we introduce the classic privacy technique Randomized Response (\cite{rr1965, Wang16}) which privatizes a variable by 'flipping' its labels based on some probability. We propose that measures of group fairness and downstream model utility can be controlled by flipping the response variable in relation to a sensitive attribute. Here, preserving model utility can be thought of as minimizing the probability the label is flipped subject to a fairness constraint that seeks to flip labels to make a training set more fair. To derive this fairness constraint, we use fair group thresholding results from recent work on Fair Bayes-Optimal Classification \cite{zeng2022bayesoptimal, zeng2022fair} which allows for fairness to be exactly controlled for. Finally, we find the solution to these optimal flipping probabilities and perturb the response variable with the corresponding randomized response mechanism, finding that downstream models trained on this perturbed variable achieve good utility at various fairness settings.


Our contributions are thus summarized as follows:
\begin{itemize}
    \item We show that a response variable can be made to satisfy many measures of group fairness at any disparity level, proposing a pre-processing method we call Fair Randomized Response (FairRR). 
    \item We extend previous theoretical results from the in-processing to the pre-processing group fairness domain.
    \item We demonstrate that classifiers trained on modified data from FairRR demonstrate excellent utility and fairness results. 
\end{itemize}

\section{PRELIMINARIES}

\subsection{Fairness}
 To introduce fair algorithmic design, we consider credit lending as an example, where it is essential to ensure lending decisions are fair in order to comply with legal requirements. This can be formulated as a fair classification problem, where two types of features are observed for potential creditors: standard features $X \in \mathcal{X}$ such as income and education, and protected (or sensitive) features $A \in \mathcal{A}$ such as gender and race. The objective is to predict the label $Y \in \{0,1\}$, if a creditor were to default on a loan, accurately and fairly with respect to $A$.  Throughout this paper, we set the sensitive feature of $A=1$  and $A=0$ respectively be some privileged and the unprivileged groups. In this way, we can split the population into four parts: the positive privileged (PP) group ($A=1$, $Y=1$), the positive unprivileged (PN) group ($A=0$, $Y=1$), the negative privileged (NP) group ($A=1$, $Y=0$),  and the negative unprivileged (NN) group ($A=0$, $Y=0$).


Researchers have proposed multiple group fairness measures for the fair classification setting. Generally, these measures depend on the constraints imposed on the joint distribution of $A$, $Y$, and a classifier's prediction $\widehat{Y}$. Common fairness measures include:
\begin{definition}[Demographic Parity]
  A prediction $\widehat{Y}$ satisfies demographic parity if it achieves the same acceptance rate  among  protected groups:
   $\Pb( \widehat{Y}  = 1|A=1) =\Pb( \widehat{Y}  = 1|A=0)  .$
\end{definition}

  \begin{definition}[Equality of Opportunity]
  A prediction $\widehat{Y}$ satisfies demographic parity if it achieves the same true positive rate  among  protected groups:
   $\Pb( \widehat{Y}  = 1|A=1,Y=1) =\Pb( \widehat{Y}  = 1|A=0,Y=1)  .$
  \end{definition}

  \begin{definition}[Predictive Equality]
 A prediction $\widehat{Y}$ satisfies predictive equality   if it achieves the same false positive rate among protected groups:
   $\Pb( \widehat{Y}  = 1|A=1,Y=0) =\Pb( \widehat{Y}  = 1|A=0,Y=0)  .$
  \end{definition}
Essentially, these notions of fairness prohibit significant mistreatment of one group over another. When the equalities holds in the aforementioned definitions, the fairness constraint enforces identical treatment among protected groups, referred to as perfect fairness. 

In practice however, a relaxed or approximate versions of these notions could be preferred as perfect fairness may require a large sacrifice of accuracy or may not be possible. This
means that instead of demanding identical treatment, we require that there should not be a significant difference in the model decisions between the two groups. Here, the disparity or unfairness of a classifier can be easily quantified by the difference between the groups. Specifically, we use $\textup{DDP}$, $\textup{DEO}$ and $\textup{DPE}$ to measure the degree of violating demographic parity, equality of opportunity, predictive equality, respectively:
\begin{align}\label{disparity_definitions}
  \begin{aligned}
    \textup{DDP}(f) &= \Pb(\widehat{Y} = 1|A = 1) -\\
    &\qquad \qquad \Pb(\widehat{Y}  = 1|A = 0)\\ \nonumber
    \textup{DEO}(f) &= \Pb(\widehat{Y} = 1|A = 1,Y=1) -\\
    &\qquad \qquad \Pb(\widehat{Y}  = 1|A = 0,Y=1) 
  \end{aligned}\\
  \begin{aligned}
    \textup{DPE}(f) &= \Pb(\widehat{Y} = 1|A = 1, Y=0) -\\
    &\qquad \qquad \Pb(\widehat{Y} = 1|A = 0,Y=0) 
  \end{aligned}
\end{align}




\subsection{Fair Bayes Optimal Classifiers under Demographic Parity}
In classification problems, the prediction  $\widehat{Y}$ is often determined by a classifier $f$ that indicates  the probability of predicting $\widehat{Y}=1$ when observing $X=x$ and $A=a$. Specifically,
  a classifier is a measurable function
  $f:\mathcal{X}\times\{0,1\} \to [0,1]$ and $Y\mid X\sim \Bern(f(X))$, with $\Bern(p)$  the Bernoulli distribution with success probability $p$. We denote by $\widehat{Y}_f$ the prediction induced by the classifier $f$ and we call $f$ is fair if its induced prediction $\widehat{f}$ satisfies the fairness constraints.
Among all fair classifiers, the Bayes optimal classifier serves as a critical theoretical benchmark, as it establishes the highest achievable accuracy for a given fairness constraint and serves as the theoretical objective that various algorithms aim to estimate.
Throughout, we will use $\textup{D}(f)$ to denote some level of disparity
 from \ref{disparity_definitions}, depending on the context.  We 
denote by $\mathcal{F}_{\delta}$ the set of measurable functions satisfying the \emph{$\delta$-parity} constraint

\begin{equation}
    \mathcal{F}_{\delta}=\{f \in \mathcal{F}: |\textup{D}(f)|\le \delta\}. \nonumber
\end{equation}

A $\delta$-fair Bayes-optimal classifier is defined as
\begin{equation}
    f_{\delta}^\star\in \underset{f\in\mathcal{F}_{\delta}}{\argmin}\, R(f)\text{ with } \ \  R(f):= \Pb\left( Y\neq \widehat{Y}_f\right). \nonumber
\end{equation}

\cite{zeng2022bayesoptimal} and \cite{zeng2022fair} studied the explicit form of fair Bayes-optimal classifiers. They found that, for many fairness metrics, the fair Bayes-optimal classifiers are group-wise thresholding rules with adjusted thresholds. Specifically, the
standard Bayes-optimal classifiers $f^*: \mathcal{X} \times \{0, 1\} \rightarrow [0, 1]$ of the form:			
$f^*(x,a)  = I\left( \eta_a(x) > {1}/{2}\right) $
can be modified to satisfy group fairness measures:
			
\begin{equation}\label{eq:FBOC}
    f_\delta^{\star}(x, a)=I\left(\eta_a(x)>\frac{1+(2a-1)T_a(t^\star_\delta)}{2}\right)
\end{equation}
Here, $(x, a) \in \mathcal{X} \times \{0, 1\}$, $\eta_a(x)=\Pb(Y=1|, A=a, X=x)$. $T_1(\cdot): \mathbb{R}\to [-1,1]$ and $T_0(\cdot): \mathbb{R}\to [-1,1]$ are two monotone non-decreasing functions with $T_1(0)=T_0(0)=0$ that are decided by the fairness metric and group-wise probabilities. In particular, with $p_{ay}=\Pb(A=a,Y=y), (a,y)\in\{0,1\}^2$, we have $T_a(t) = t/(p_{a1}+p_{a0})$ for demographic parity,  $T_a(t) = t/[2p_{a1}-(2a-1)t]$ for equality of opportunity, and $T_a(t) = t/[2p_{a0} +(2a-1)t]$ for predictive equality. 

The parameter $t_{\delta}^{\star}$ is decided by the disparity level $\delta$ where for a given $t$ in a proper range, the classifier $f_t(x, a)=I\left(\eta_a(x)>\frac{1+(2a-1)T_a(t)}{2}\right)$ is a fair Bayes-optimal classifier for a certain disparity level $\delta_t$. In particular, the disparity level $D(t) = D(f_t)$ is a monotone non-increasing function of $t$.
In other words, $t_{\delta}^{\star}$ can be thought of as a term that balances the fairness-accuracy tradeoff of the fair Bayes-optimal classifier. Details on estimating $t_{\delta}^{\star}$ can be found in the next section, but in practice it can also be treated as a hyperparameter to control for disparity.

\subsection{Design Matrices in Randomized Response}

Randomized Response was first proposed by \cite{rr1965} to preserve the privacy of survey respondents' answers when asked sensitive questions and is a classic privacy technique. To start, suppose $n$ individuals each have a response for some sensitive binary attribute $Y$, $y_i \in {0, 1}$. Each individual wishes to preserve the privacy of their response and so they send to an untrusted server a modified version of $y_i$ in which the label is flipped to $\widetilde{y}_i$ by some probability.
The probabilities in which $y_i$ is flipped are determined by a design matrix which in the binary case can be written as:

\begin{equation}\label{design_matrix}
\textbf{P} =
\begin{bmatrix}
P(\widetilde{Y} = 1 | Y = 1) & P(\widetilde{Y} = 1 | Y = 0)\\
P(\widetilde{Y} = 0 | Y = 1) & P(\widetilde{Y} = 0 | Y = 0)
\end{bmatrix}
\end{equation}

To anonymize $Y$ across a second binary variable $A \in \{0,1\}$, we can rewrite \ref{design_matrix} to consist of separate design matrices:

\begin{equation}\nonumber
\begin{aligned}
    &\textbf{P}_1 =  \\
    &\begin{bmatrix}
    P(\widetilde{Y} = 1 | A=1, Y = 1) & P(\widetilde{Y} = 1 | A=1, Y = 0)\\
    P(\widetilde{Y} = 0 | A=1, Y = 1) & P(\widetilde{Y} = 0 | A=1, Y = 0)
    \end{bmatrix}
\end{aligned}
\end{equation}

and
\begin{equation}\nonumber
\begin{aligned}
&\textbf{P}_0 = \\
    &\begin{bmatrix}
    P(\widetilde{Y} = 1 |A=0, Y = 1) & P(\widetilde{Y} = 1 |A=0, Y = 0)\\
    P(\widetilde{Y} = 0 | A=0,Y = 1) & P(\widetilde{Y} = 0 |A=0, Y = 0)
    \end{bmatrix}
\end{aligned}
\end{equation}

Since the columns for each matrix must sum to 1, $\textbf{P}_1$ and $\textbf{P}_0$ can be expressed as the randomization mechanism $\mathcal{R}_{(\theta_{11},\theta_{10},\theta_{01},\theta_{00})}$ where $\theta_{ay} = \Pb(\widetilde{Y} = y| A = a, Y = y)$:\\
$$
\textbf{P}_1 = 
\begin{bmatrix}

\theta_{11} & 1-\theta_{10}\\
1-\theta_{11} & \theta_{10}

\end{bmatrix}
$$
and
$$
\textbf{P}_0 = 
\begin{bmatrix}

\theta_{01} & 1-\theta_{00}\\
1-\theta_{01} & \theta_{00}

\end{bmatrix}
$$

\section{METHOD}
\subsection{Overview}

We therefore have the preliminaries to begin developing a pre-processing method to perturb $Y$ to be fair. Here, the goal is to find the randomization mechanism $\mathcal{R}_{(\theta_{11},\theta_{10},\theta_{01},\theta_{00})}$ that maximizes downstream model utility subject to fairness constraints. The design matrix for the  best randomization mechanism can be easily found before then being applied to the training dataset. After this application, a final classifier can then be fit for the original $X$ and now perturbed label variable $\widetilde{Y}$. 

To start, we propose that the best $\mathcal{R}_{(\theta_{11},\theta_{10},\theta_{01},\theta_{00})}$ from solely a utility perspective would be the one that does not flip $Y$ at all as it would not inject any noise into the training dataset. Thus, we wish to maximize $\Pb(\widetilde{Y}=Y)$ or:

\begin{equation*}\label{equ:objective}
  \max p_{11}\theta_{11} + p_{10}\theta_{10} +  p_{01}\theta_{01} +p_{00}\theta_{00}   
\end{equation*}

where $\frac{1}{2} \leq \theta_{11}, \theta_{10}, \theta_{01}, \theta_{00} \leq 1$. We will show that common group definitions of fairness can be written as linear equality constraints for this function.

\subsection{Fairness through Randomized Response}
\subsubsection{Randomized Response and the Fair Bayes-Optimal Classifier}
To illustrate how fairness can be achieved by randomized response, we first consider the Bayes-Optimal classifier with no protected attribute. Specifically, let $\mathcal{R}_{\theta_1,\theta_2}$ denote the randomization mechanism as follows: $\widetilde{Y}=\mathcal{R}_{\theta_1,\theta_0}(Y)$ has the property that $\Pb(\widetilde{Y}=Y) =\theta_1$ for $Y=1$ and $\theta_0$ for $Y=0$ for any $(\theta_1,\theta_0)\in [1/2,1]\times[1/2,1]$. Note that this mechanism is im-balanced if $\theta_1\neq \theta_0.$ Denote by $\eta(x)$ and  $\widetilde{\eta}(x)$ the conditional distribution of $Y$ and $\widetilde{Y}$ given $X = x$, respectively. It can be verified that:
$$\widetilde{\eta}(x)= \theta_1\eta(x) + (1 - \theta_2)(1 - \eta(x)).$$

Clearly, $\widetilde{\eta}(x)>1/2$ is equivalent to $\eta(x)>(\theta_0-1/2)/(\theta_1+\theta_0-1)$. Hence, $\mathcal{R}_{\theta_1,\theta_0}(Y)$ essentially shifts the thresholds of the decision rule when $\theta_1\neq \theta_0$.
As we discussed in section 2, the optimal fair classifiers are known to be group-wise thresholding
rules for many fairness-metrics. 
The aformentioned technical connection enables us to generate
a fair dataset through an im-balanced randomization of response. 
\begin{theorem}\label{thm:rr} Let $(X,A,Y)$ follow a distribution $\Pb$ on $\mathcal{X}\times \{0,1\}\times \{0,1\}$. 
Consider a group-wise im-balanced randomized response mechanism $\widetilde{Y}=\mathcal{R}_{\theta_{11},\theta_{10},\theta_{01},\theta_{00}}(A,Y)$ with, for $a\in\{0,1\}$,
\begin{equation}
\Pb(\widetilde{Y}=Y|A=a)=\left\{
\begin{array}{ll}
  \theta_{a1}, &  \text{ for } Y=1;\\
  1-\theta_{a0}, &  \text{ for } Y=0.\\
\end{array}\right.
\end{equation}
When the flipping probabilities satisfy:
\begin{eqnarray*}
&&(T_1(t_\delta^\star)+1)\theta_{11}+(T_1(t^\star_\delta)-1)\theta_{10}=T_1(t^\star_\delta);\\
&&(T_0(t_\delta^\star)-1)\theta_{01}+(T_0(t^\star_\delta)+1)\theta_{00}=T_0(t^\star_\delta);
\end{eqnarray*}
where $T_1(\cdot)$, $T_0(\cdot)$ and $t^\star$ are the same as in \eqref{eq:FBOC}. Denote $\widetilde{\Pb}$ as the joint distribution of $(X,A,\widetilde{Y})$. Then, the Bayes optimal classifier learned on $\widetilde{\Pb}$ is a $\delta$-fair Bayes-optimal classifier \eqref{eq:FBOC} learned on $\Pb$. 
\end{theorem}

\begin{remark}
We need to maximize $\theta_{ay} \in [1/2,1]$ to maximize the objective function \eqref{equ:objective}. As a result, we can take, when $t^\star_\delta>0$,
\begin{eqnarray}{\label{eq:solve_theta}}
(\theta_{11},\theta_{10},\theta_{01},\theta_{00}) = 
   \left(\frac{1}{1+T_1(t^\star_\delta)}, 1 ,1 , \frac{1}{1+T_0(t^\star_\delta)} \right), 
\end{eqnarray}
and,  when $t^\star_\delta<0$,
\begin{eqnarray}{\label{eq:solve_theta1}}
(\theta_{11},\theta_{10},\theta_{01},\theta_{00}) = 
   \left(1,\frac{1}{1-T_1(t^\star_\delta)},   \frac{1}{1-T_0(t^\star_\delta)},1 \right), 
\end{eqnarray}
\end{remark}
By Theorem \ref{thm:rr} we can express group fairness definitions such as Demographic Parity, Equalized Opportunity, and Predictive Equality in terms of the randomization mechanism which double as equality constraints.

\begin{definition}[Demographic Parity]\label{RR:DP}
  A randomization mechanism achieves Demographic Parity if it satisfies:
\begin{eqnarray}
&&(p_{11}+p_{10}+t_\delta^\star)\theta_{11}+(t^\star_\delta-p_{11}-p_{10})\theta_{10}=t_\delta^\star;\nonumber \\
&&(t_\delta^\star-p_{01}-p_{00})\theta_{01}+(t^\star_\delta+p_{01}+p_{00})\theta_{00}=t_\delta^\star. \nonumber
\end{eqnarray}
  \end{definition}
  
\begin{definition}[Equality of Opportunity]\label{RR:EO}
  A randomization mechanism achieves Equalized Opportunity if it satisfies:
\begin{eqnarray}
&& 2p_{11}\theta_{11}+2(t^\star_\delta- p_{11})\theta_{10}=t_\delta^\star; \nonumber\\ 
&&-2p_{01}\theta_{01}+2(t_\delta^\star+p_{01})\theta_{00}=t_\delta^\star. \nonumber
\end{eqnarray}
  \end{definition}

\begin{definition}[Predictive Equality]\label{RR:PE}
    A randomization mechanism achieves Predictive Equality if it satisfies:
\begin{eqnarray}
&&2(t^\star_\delta+p_{10})\theta_{11}-2p_{10}\theta_{10}=t_\delta^\star; \nonumber\\ 
&&2(t^\star_\delta -p_{00})\theta_{01}+2p_{00}\theta_{00}=t_\delta^\star. \nonumber\end{eqnarray}
  \end{definition}

\begin{table*}[t]
\caption{Benchmarking Results: Original vs FairRR Pre-processed Datasets}
\label{Benchmarking Results: Original vs FairRR Preprocessed Datasets}
\centering
\begin{tabular}{@{\extracolsep{4pt}}lcccccc}
\multicolumn{6}{c}{}\\
\multicolumn{6}{c}{Panel A: Original Datasets}\\
\toprule  
{} &    \multicolumn{5}{c}{Metrics} \\
\cmidrule{2-6} 
Datasets &   \multicolumn{1}{c}{Acc} & $f_1$ & DDP & DEO & DPE\\ 
\cmidrule{1-1}\cmidrule{2-6}
Adult       &  0.841  &  0.620  &  0.188  &  0.184  &  0.086  \\ 
            & (0.003) & (0.007) & (0.006) & (0.026) & (0.005) \\ 
\cmidrule{1-1}\cmidrule{2-6}
COMPAS      &  0.676  &  0.632  &  0.283  &  0.313  &  0.186  \\ 
            & (0.015) & (0.016) & (0.031) & (0.052) & (0.035) \\
\cmidrule{1-1}\cmidrule{2-6}
Law School  &  0.787  &  0.499  &  0.060  &  0.084  &  0.024  \\ 
            & (0.003) & (0.005) & (0.005) & (0.015) & (0.004) \\ 
\bottomrule
\end{tabular}
\begin{tabular}{@{\extracolsep{4pt}}lccccccccc}
\multicolumn{10}{c}{ }\\
\multicolumn{10}{c}{Panel B: Fair Randomized Response}\\
\toprule  
{} &   \multicolumn{9}{c}{Fairness Criteria} \\
\cmidrule{2-10}
{}  & \multicolumn{3}{c}{Demographic Parity} & \multicolumn{3}{c}{Equality of Opportunity} &\multicolumn{3}{c}{Predictive Equality}\\
\cmidrule{2-4}\cmidrule{5-7}\cmidrule{8-10}
{} &    \multicolumn{3}{c}{Metrics} &    \multicolumn{3}{c}{Metrics} &    \multicolumn{3}{c}{Metrics} \\
\cmidrule{2-4}\cmidrule{5-7}\cmidrule{8-10}
Datasets     &   Acc   &  $f_1$  &   DDP   &   Acc   &  $f_1$  &   DEO   &   Acc   &  $f_1$  &   DPE   \\ 
\cmidrule{1-1}\cmidrule{2-4}\cmidrule{5-7}\cmidrule{8-10}
Adult          &  0.820 & 0.534 & 0.007 & 0.839 & 0.608 & 0.024 & 0.829 & 0.563 & 0.005\\
            & (0.004) & (0.009) & (0.005) & (0.003) & (0.007) & (0.02) & (0.004) & (0.009) & (0.004) \\ 
\cmidrule{1-1}\cmidrule{2-4}\cmidrule{5-7}\cmidrule{8-10}
COMPAS      &  0.660 & 0.608 & 0.027 & 0.661 & 0.610 & 0.046 & 0.667 & 0.614 & 0.031  \\ 
            & (0.015) & (0.017) & (0.019) & (0.014) & (0.016) & (0.037) & (0.014) & (0.016) & (0.024) \\ 
\cmidrule{1-1}\cmidrule{2-4}\cmidrule{5-7}\cmidrule{8-10}
Law School  & 0.785 & 0.486 & 0.006 & 0.785 & 0.489 & 0.015 & 0.786 & 0.493 & 0.004  \\
            & (0.003) & (0.005) & (0.004) & (0.003) & (0.005) & (0.011) & (0.003) & (0.005) & (0.004) \\ 
\bottomrule
\end{tabular}
\end{table*}


\subsubsection{FairRR: a Randomized Response Mechanism for Fair Classification}
In this section, we propose the Randomized Response Mechanism that 
removes the discrimination from the training dataset. Based on the aformentioned theory, we are able to derive the optimal fair flipping probabilities as long as we estimate $p_{ay}, (a,y)\in\{0,1\}^2$ and $t^\star_\delta$ from the training data. $p_{ay}$ can be estimated directly by using its empirical estimator and  $t^\star_\delta$ can be conveniently estimated using bisection methods due to its monotonic relationship with the decision disparity. 

Here, we set $t_{min}=\inf_t: \{|T_a(t)|\le 1 \text{ for }a \in \{0,1\}, \}$ and $t_{max} = \sup_t: \{|T_a(t)|\le 1 \text{ for }a \in \{0,1\}, \}$. In each iteration, we update $t = (t_{max}+t_{min})/2$ and calculate the flipping probabilities as referenced in \eqref{eq:solve_theta} and \eqref{eq:solve_theta1}. Then, classifier $\hat{f}_t$ is learned from $(X,A,\widetilde{Y})$ with $\widetilde{Y}= \mathcal{R}_{\theta_{11},\theta_{10},\theta_{01},\theta_{00}}(Y)$. If the disparity level of $\hat{f}_t$ is greater than the pre-specified disparity level, we set $t_{min} = t_{mid}$ iterate until $t^\star_\delta $ is found.

Thus, with $p_{ay}$ and $t^\star_\delta$ estimated, the optimal $(\theta_{11},\theta_{10},\theta_{01},\theta_{00})$ can be solved for using \eqref{eq:solve_theta}, \eqref{eq:solve_theta1} and a corresponding group fairness definition, which maximizes \ref{equ:objective} subject to the constraints of either \ref{RR:DP}, \ref{RR:EO}, or \ref{RR:PE}. With this randomization mechanism, the values in the privileged group $A = 1$ are randomly flipped from $Y = 1$ to $Y = 0$ and values in the unprivileged group $A = 0$ are randomly flipped from $Y = 0$ to $Y = 1$ such that a new perturbed response variable $\widetilde{Y}$ is created. Any classifier can then be fit to the original data $X$ with perturbed $\widetilde{Y}$. 

The time complexity of this method is dependent on which classifier is chosen for estimating $t^\star_\delta$ in the aforementioned bisection method. Here, a classifier has to be iteratively trained and evaluated to find the desired $t^\star_\delta$. In practice, we find that this takes relatively few iterations. In the case of using logistic regression with the LBFGS solver for example, the training complexity is O($p\times m$) where $p$ is the number of parameters and $m$ is the number of memory corrections (\cite{bigO}). The evaluation of each iteration simplifies to O($n\times p$) where $n$ is the size of the evaluation set. Once $t^\star_\delta$ is estimated, the perturbation of $Y$ is an O($N$) process where $N$ is the sample size of all data to be perturbed. The overall time complexity is thus dependent on $n$, $m$, $p$, and $N$ as to what the final complexity reduces to.

\begin{table*}[t]
\caption{Benchmarking Results: Pre-processing Methods}
\label{Benchmarking Results: Preprocessing Methods}
\centering
\begin{tabular}{@{\extracolsep{4pt}}lllcccc}
\toprule   
{} & {} &  \multicolumn{5}{c}{Methods} \\
\cmidrule{3-7} 
Datasets & Metrics & Original & FairRR & TabFairGan & FS & FAWOS\\ 
\midrule
Adult & Acc & 0.841 & 0.820 & 0.804 & 0.836 & 0.786\\ 
      &     & (0.003) & (0.004) & (0.008) & (0.003) & (0.004) \\ 
      & DDP & 0.188 & 0.007 & 0.023 & 0.091 & 0.008 \\ 
      &     & (0.006) & (0.005) & (0.024) & (0.008) & (0.006) \\
\midrule
COMPAS & Acc & 0.676 & 0.660 & 0.631 & 0.659 & 0.632\\ 
        &     & (0.015) & (0.015) & (0.034) & (0.014) & (0.015) \\
        & DDP & 0.283 & 0.027 & 0.150 & 0.033 & 0.022\\
        &     & (0.031) & (0.019) & (0.110) & (0.026) & (0.017) \\
\midrule
Law School & Acc & 0.787 & 0.785 & 0.774 & 0.784 & 0.782\\
           &     & (0.003) & (0.003) & (0.030) & (0.003) & (0.003)\\
           & DDP & 0.060 & 0.006 & 0.060 & 0.006 & 0.006 \\
           &     & (0.005) & (0.004) & (0.153) & (0.004) & (0.004)\\ 
\bottomrule
\end{tabular}
\end{table*}

\begin{table*}[t]
\caption{Direct Control on Pre-specified Disparity Levels ($\delta$)}
\label{Benchmarking Results: Direct Control}
\centering
\begin{tabular}{@{\extracolsep{4pt}}lllcccc}
\toprule   
Datasets & \multicolumn{6}{c}{Metrics}  \\ 
\midrule
Adult & $\delta$ &   0.000  &  0.040  &  0.080  &  0.120  &  0.160 \\
\cmidrule{2-7} 
      & DDP &   0.007  &  0.040  &  0.081  &  0.121  &  0.161  \\
      &     & (0.005) & (0.008) & (0.008) & (0.008) & (0.007) \\
      &  Acc &   0.820  &  0.826  &  0.833  &  0.838  &  0.841  \\
      &     & (0.004) & (0.004) & (0.003) & (0.003) & (0.003) \\
\midrule
COMPAS & $\delta$ &    0.000  &  0.060  &  0.120  &  0.180  &  0.240  \\
\cmidrule{2-7} 
      & DDP &    0.027  &  0.062  &  0.123  &  0.182  &  0.239  \\
      &     &  (0.019) & (0.030) & (0.032) & (0.030) & (0.032) \\
      &  Acc &    0.660  &  0.665  &  0.669  &  0.674  &  0.676  \\
      &     &  (0.015) & (0.014) & (0.014) & (0.015) & (0.015) \\
\midrule
Law School & $\delta$ &   0.000  &  0.012  &  0.024  &  0.036  &  0.048  \\
\cmidrule{2-7} 
      &  DDP &   0.006  &  0.013  &  0.025  &  0.036  &  0.049  \\
      &     &  (0.004) & (0.006) & (0.007) & (0.006) & (0.006) \\
      & Acc &     0.785  &  0.785  &  0.786  &  0.786  &  0.786  \\
      &     &  (0.003) & (0.003) & (0.003) & (0.003) & (0.003) \\
\bottomrule
\end{tabular}
\end{table*}

\section{EXPERIMENTS}

\subsection{Empirical Data Analysis}

{\bf Datasets:} We test FairRR on three  benchmark  datasets for fair classification: Adult \cite{Dua2019}, COMPAS \cite{JJSL2016} and Law School \cite{wightman1998lsac}.

\begin{itemize}
\item {\it Adult:} The target variable $Y$ is whether the income of an individual is more than \$50,000.  Age, marriage status, education level and other related variables are included in $X$, and the protected attribute $A$ refers to gender. 

\item {\it COMPAS:} In the COMPAS dataset, the target is to predict recidivism. Here $Y$ indicates whether or not a criminal will reoffend, while $X$ includes  prior criminal records, age and an indicator of misdemeanor. 
The protected attribute $A$  is the race of an individual, ``white-vs-non-white''.

\item {\it Law School:} The task of interest in Law School data set is to predict whether an applicant gets an
admission from a law school based on common features include LSAT score and undergraduate GPA.  The protected attribute $A$  is the race of the individual: “white-vs-non-white”
\end{itemize}

{\bf Compared algorithms:} In addition to FairRR, we also consider several benchmark methods in our experiments. As FairRR is a pre-processing method, we only include other pre-processing methods for comparison. Specifically, we consider the following:

\begin{itemize}


\item (1) Fair Sampling

Fair Sampling \cite{KC2012} is a method based on adjusting the size of PP, PN, NP and NN groups. Its idea is to apply over/down sampling such that the label on the training data is independent of the sensitive attribute. Specifically, size of group PP, PN, NP and NN after sampling are:
\begin{equation}
    n_{ay} = \frac{(n_{a1} + n_{a0})(n_{1y} + n_{0y})}{n_{11} +  n_{10} + n_{01}+ n_{00} }\nonumber
\end{equation}

\item (2) FAWOS

FAWOS \cite{FAWOS} is another sampling method for fairness proposed recently. Unlike fair sampling that adjust the sizes of all four groups, FAWOS only applies SMOTE (\cite{chawla2002smote}, a popular oversampling method for unbalanced classification problem) to over-sample the points in the NN group where the number of points generated is:
\begin{equation}
   N = \alpha \times \left(\frac{n_{11} n_{00}}{n_{10}} - n_{01}\right) \nonumber
\end{equation}

\item (3) TabFairGAN 

TabFairGAN \cite{rajabi2021tabfairgan} is a fair synthetic generation method based on the framework of generative adversarial network which adds a fairness penalty term to the generator loss of a standard WGAN model. 
Specifically, the fairness penalty is equal to the demographic parity
 of the generated data squared.

\end{itemize}

\textbf{Experimental Setting:}
The goal of fair classification is to learn a classifier with the highest model utility, subject to some fairness constraint. Thus to test and benchmark FairRR we first apply each aforementioned fair pre-processing algorithm to each training dataset. A logistic regression classifier is then learned on the returned de-biased training dataset where it is then evaluated based on the average accuracy, $f_1$ score and disparity levels over 100 random 80:20 train/test splits. All model hyperparameters are left as the scikit-learn defaults for reproduciblity. The standard deviations of these metrics are also reported. All training and evaluations were processed using an Apple M1 CPU.

\subsection{Results}
We first evaluate the performance of FairRR controlling for either Demographic Parity, Equalized Opportunity, or Predictive Equality. We present the simulation
results in Table \ref{Benchmarking Results: Original vs FairRR Preprocessed Datasets}. We observe that FairRR significantly controls for disparity across each fairness metric while seeing minimal decreases of model utility measured by accuracy and $f_1$ score. We then benchmark FairRR with other existing pre-processing methods. Here, only demographic parity is considered as it is the only common fairness metric supported across all pre-processing methods. We present these benchmarking results in Table \ref{Benchmarking Results: Preprocessing Methods}. 

\begin{figure}
    \centering
        \includegraphics[width=.8\linewidth]{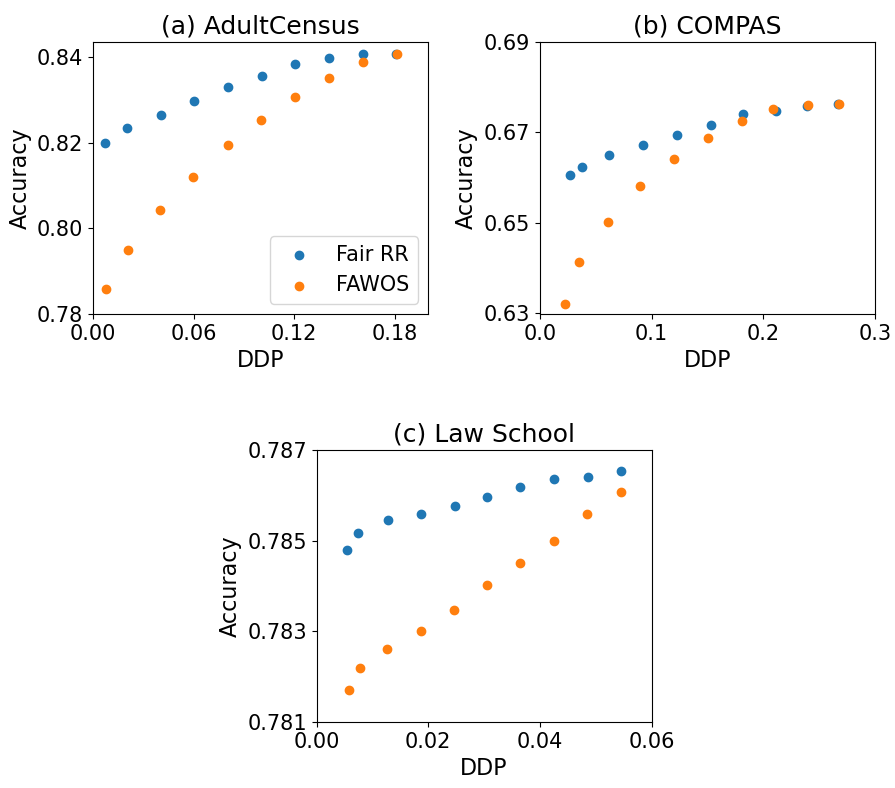}
        \caption{Logistic Regression Accuracy/ Disparity Trade-offs: FairRR and FAWOS comparison across datasets.}
        \label{fig:subfig1}
    \end{figure}

\begin{figure}
    \centering
        \includegraphics[width=.7\linewidth]{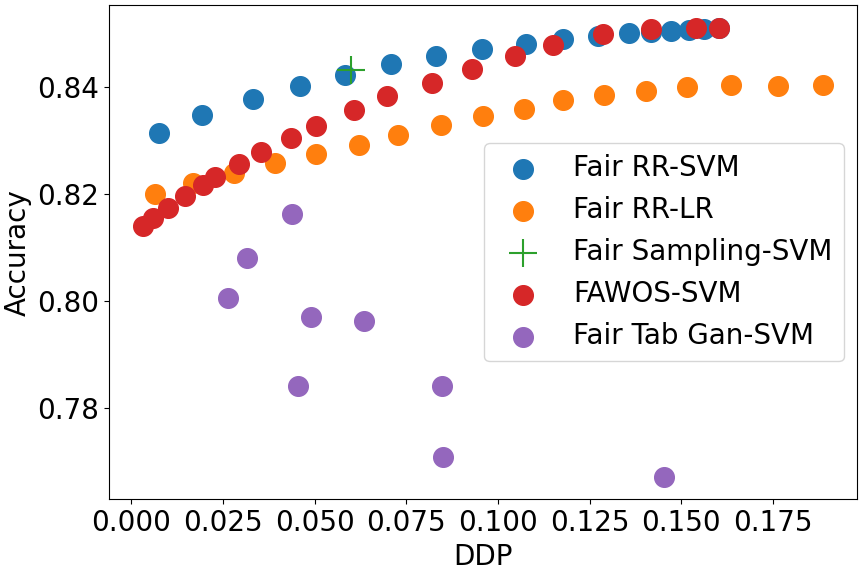}
        \caption{Accuracy/ Disparity Pareto Curves of various pre-processing algorithms on the Adult dataset evaluated with Support Vector Machines.}
        \label{fig:subfig2}
    \end{figure}

Finally, we showcase the ability of FairRR to control for specific levels of disparity. In Table \ref{Benchmarking Results: Direct Control}, FairRR was set to control for disparity at the quintiles between perfect demographic parity and the DDP level of the original dataset. The corresponding average DDP values in the final logistic regression and corresponding accuracies with standard deviations over 100 random seeds are reported. Figure \ref{fig:subfig1} plots this experiment to highlight the accuracy/ disparity trade-off, comparing FairRR to FAWOS at these quintiles of controlled-for disparity. Figure \ref{fig:subfig2} showcases the Pareto Curves of FairRR, Fair Sampling, FAWOS, and FairTabGAN when an SVM is trained on pre-processed data from the Adult Dataset. 

\section{DISCUSSION}

Overall, FairRR achieves favorable or comparable-to-the-leader accuracy and disparity scores across the three benchmarking datasets. FairRR effectively maintains model utility while enforcing small amounts of disparity, regardless of the chosen group fairness definitions. One surprising result was the stability of the algorithm. One potential downside to FairRR could be with it randomly flipping labels the effectiveness could vary widely depending on the random seed. With low standard deviations across evaluation metrics though, FairRR proves to be also be robust. One interesting finding is that FairRR, Fair Sampling (FS), and FAWOS generally performed better than TabFairGan. We suspect this is because TabFairGan learns both $X$ and $y$, which has advantages for applications such as privacy, but likely makes it weaker for pure fair classification tasks where FairRR and the over/under sampling strategies in FS and FAWOS perturb the feature space less.

FairRR also favorably controls for exact levels of disparity, an added benefit in applications where perfect group fairness is impractical or not needed. Table \ref{Benchmarking Results: Direct Control} shows empirically that disparity can be set to a level a-priori to model training and the downstream model will have that final level of disparity. Similarly, the trade-off between accuracy and disparity is better than competing methods that have this feature. With FAWOS conveniently allowing for the control of disparity we compare it with FairRR in Figure \ref{fig:subfig1}, showing that FairRR has a preferable utility curve to FAWOS in that at nearly all levels of disparity, the model trained with FairRR-processed data has better accuracy. This is further shown in Figure \ref{fig:subfig2} which evaluates all competing methods on the Adult dataset with Support Vector Machines. Here, FairRR dominated the Pareto Frontier, noting that Fair Sampling does not allow for disparity control.

Another component investigated was the corresponding privacy offered by FairRR. As Randomized Response was first introduced as a privacy method, a natural extension of FairRR is to investigate the relationship between its utility/ fairness trade-off and the additional privacy it provides. This proves to be technically challenging. While Randomized Response is shown to satisfy ($\epsilon, \delta$)- Label Differential Privacy (\cite{Wang16}, \cite{shirong2023binary}), the added fairness component of FairRR complicates a typical privacy analysis as it makes the privacy mechanism no longer independent of the data it is privatising. This is highlighted in the estimation of $t_{\delta}^{\star}$ where the design matrix is explicitly calculated based off of the disparity level in the original dataset. While $\widetilde{Y}$ is more private than $Y$, it remains unsolved how to quantify exactly how much more private it is in the context of some privacy budget. However, in application an advantage of pre-processing is that other pre-processing techniques can also be applied to the training data and in the context of privacy, FairRR is well-suited to be used in conjunction with other privacy mechanisms such as Laplacian and Exponential Noise (\cite{jain_differential_2018}).

\section{CONCLUSION}

FairRR can be an excellent choice achieving group fairness in that it is a downstream model agnostic, efficient, and theory motivated algorithm that supports most group fairness definitions. In benchmarking, it performs comparably or better than other choices for pre-processing algorithms and additionally connects previous fair statistical learning theory to the pre-processing domain.

There are a variety of future research opportunities with FairRR. For starters, this paper only addresses the single binary sensitive attribute, single binary outcome problem formulation of fair classification. We believe that FairRR could be generalized to work in settings where multiple sensitive attributes are needed. Another interesting line of work is studying FairRR in the context of privacy, what Randomized Response was initially designed for. While this is technically challenging, we believe that extensions on FairRR could help shed light into the theoretical trade-offs between fairness and privacy. Lastly, we suspect there are a variety of additional mechanisms outside of randomized response to further apply the idea of pre-processing or post-processing data based on the Fair Optimal Bayes thresholding to achieve group fairness.

\subsubsection*{Acknowledgements}
We thank the anonymous reviewers for their helpful comments and suggestions. This work was partially supported by the JP Morgan Chase Faculty Research Award, NSF -- CNS (2247795), and the Office of Naval Research (ONR N00014-22-1-2680). This work is also partially supported by the National Natural Science Foundation of China, No. 72033002.

\bibliography{RR_Bibliography}

\end{document}


%

%

\onecolumn
\aistatstitle{FairRR: Pre-Processing for Group Fairness through Randomized Response \\
Supplementary Materials}
\chead{\textbf{FairRR: Pre-Processing for Group Fairness through Randomized Response}}

\section{MISSING PROOFS}

\subsection{Proof of Theorem 3.1}
For $x\in \mathcal{X}$  and $a\in\mathcal\{0,1\}$, we denote $\eta_a(x)= \mathbb{P}\left( Y=1|A=a,X=x\right)$ and $\widetilde\eta_a(x)= \mathbb{P}\left( \widetilde{Y}=1|A=a,X=x\right)$. By construction, we have,
$$
\left\{
\begin{array}{l}
\mathbb{P}\left( \widetilde{Y}=1|A=a,Y=1\right) =\theta_{a1}\\
\mathbb{P}\left( \widetilde{Y}=1|A=a,Y=0\right) =1-\theta_{a0}.
\end{array}
\right.
$$
It follows that,
\begin{align*}
\widetilde\eta_a(x) =&\mathbb{P}\left( \widetilde{Y}=1|A=a,X=x\right)\\
=&\mathbb{P}\left( \widetilde{Y}=1,Y=1|A=a,X=x\right) +\mathbb{P}\left( \widetilde{Y}=1,Y=0|A=a,X=x\right)\\
=&\mathbb{P}\left( \widetilde{Y}=1|Y=1,A=a,X=x\right)\cdot\mathbb{P}\left( {Y}=1|A=a,X=x\right) \\
&+\mathbb{P}\left( \widetilde{Y}=1|Y=0,A=a,X=x\right)\cdot\mathbb{P}\left( {Y}=0|A=a,X=x\right) \\
=&\mathbb{P}\left( \widetilde{Y}=1|A=a,Y=1\right)\cdot\mathbb{P}\left( {Y}=1|A=a,X=x\right) \\
&+\mathbb{P}\left( \widetilde{Y}=1|A=a,Y=0\right)\cdot\left( 1-\mathbb{P}\left( {Y}=1|A=a,X=x\right)\right) \\
=&  \theta_{a1}\eta_a(x)+(1- \theta_{a0})(1-\eta_a(x)).
\end{align*}
On one hand, by \cite{DevroyeGL96}, the Bayes optimal classifier learned on ${\mathbb{P}}$ is
$$\widetilde{f}^\star(x,a) = I\left(\widetilde\eta_a(x)>\frac12\right) =  I\left(\theta_{a1}\eta_a(x)+(1- \theta_{a0})(1-\eta_a(x))>\frac12\right) ,$$
which is equivalent to,
\begin{equation}\label{eq:FBOC1}
\widetilde{f}^\star(x,a)=
I\left( \eta_a(x)> \frac12 + \frac{\theta_{a0}-\theta_{a1}}{2\theta_{a1}+2\theta_{a0}-2}\right) .\end{equation}

On the other hand, by \cite{zeng2022bayesoptimal}, for many group fairness metrics, the fair Bayes optimal classifier learned on $\mathbb{P}$ can be written as,
\begin{equation}\label{eq:FBOC2}
    f_\delta^{\star}(x, a)=I\left(\eta_a(x)>\frac{1+(2a-1)T_a(t^\star_\delta)}{2}\right),
\end{equation}
where $T_1(\cdot):\mathbb{R}\to [-1,1]$ and $T_0(\cdot): \mathbb{R}\to [-1,1]$ are two monotone non-decreasing functions with $T_1(0)=T_0(0)=0$ that are decided by the fairness metric and group-wise probabilities. 

Clearly, \eqref{eq:FBOC1} and  \eqref{eq:FBOC2} are the same if and only if, for $a\in\{0,1\}$,
$$\frac{\theta_{a0}-\theta_{a1}}{2\theta_{a1}+2\theta_{a0}-2} =\frac{(2a-1)T_a(t^\star_\delta)}{2},$$
or equivalently,
\begin{equation}\label{eq:relationship}
\begin{array}{l}
(T_1(t_\delta^\star)+1)\theta_{11}+(T_1(t^\star_\delta)-1)\theta_{10}=T_1(t^\star_\delta);\\
(T_0(t_\delta^\star)-1)\theta_{01}+(T_0(t^\star_\delta)+1)\theta_{00}=T_0(t^\star_\delta).
\end{array}
\end{equation}

In the following, we derive the flipping probabilities for demographic parity, equality of opportunity and predictive equality in order. For $(a,y)\in\{0,1\}^2$, denote $p_{ay}= \mathbb{P}(A=a,Y=y)$. According to \cite{zeng2022bayesoptimal}, $T_1(\cdot)$ and $T_0(\cdot)$ take forms, for $t\in\mathbb{R}$,
\begin{equation*}
 \left( T_1(t),T_0(t)\right) =\left\{
 \begin{array}{ll}
   \left(\frac{t}{p_{11}+p_{10}},  \frac{t}{p_{01}+p_{00}}\right),   &  \text{ for demographic parity};\\
   \left(\frac{t}{2p_{11}-t} ,   \frac{t}{2p_{01}+t}\right),   &  \text{ for equality of opportunity};\\
      \left(\frac{t}{2p_{10}+t} ,\frac{t}{2p_{00}-t}\right),   &  \text{ for predictive equality 
}.
 \end{array}
 \right.
\end{equation*}
Thus, \eqref{eq:relationship} can be written as,
\begin{itemize}
    \item for demographic parity, 
\begin{eqnarray*}
&&(p_{11}+p_{10}+t_\delta^\star)\theta_{11}+(t^\star_\delta-p_{11}-p_{10})\theta_{10}=t_\delta^\star;\nonumber \\
&&(t_\delta^\star-p_{01}-p_{00})\theta_{01}+(t^\star_\delta+p_{01}+p_{00})\theta_{00}=t_\delta^\star;
\end{eqnarray*}
    \item for equality of opportunity, 
\begin{eqnarray*}
&&2p_{11}\theta_{11}+2(t^\star_\delta- p_{11})\theta_{10}=t_\delta^\star;\\
&&-2p_{01}\theta_{01}+2(t_\delta^\star+p_{01})\theta_{00}=t_\delta^\star;
\end{eqnarray*}
    \item and for predictive equality, 
\begin{eqnarray*}
&&2(t^\star_\delta+p_{10})\theta_{11}-2p_{10}\theta_{10}=t_\delta^\star; \\ 
&&2(t^\star_\delta -p_{00})\theta_{01}+2p_{00}\theta_{00}=t_\delta^\star. 
\end{eqnarray*}
\end{itemize}

\bibliographystyle{plain}
\bibliography{RR_Bibliography}

\vfill